\documentclass[conference]{IEEEtran}
\IEEEoverridecommandlockouts

\usepackage{cite}
\usepackage{amsmath,amssymb,amsfonts}
\usepackage{graphicx}
\usepackage{algorithmic}
\usepackage{algorithm}
\usepackage{textcomp}
\usepackage{amsthm}
\usepackage{multicol}
\usepackage{multirow}
\usepackage{color}
\usepackage{subcaption}
\usepackage{import}
\usepackage{enumitem}

\usepackage[utf8]{inputenc} 
\usepackage[T1]{fontenc}    
\usepackage{hyperref}       
\hypersetup{
	colorlinks,
	linkcolor={black},
	citecolor={blue!80!black},
	urlcolor={blue!80!black}
}
\usepackage{url}            
\usepackage{booktabs}       
\usepackage{amsfonts}       
\usepackage{nicefrac}       
\usepackage{microtype}      
\usepackage[pdftex,dvipsnames]{xcolor}         
\usepackage{comment}
\usepackage{soul}
\usepackage{float}
\usepackage[normalem]{ulem}
\usepackage{xargs} 

\usepackage[
colorinlistoftodos,prependcaption,
textcolor=red,
linecolor=red,
bordercolor=red,
backgroundcolor=white,
]{todonotes}

\begin{document}
\title{Neural Greedy Pursuit for Feature Selection \\
\thanks{We thank the Swedish Foundation for Strategic Research for their funding to support the work.}
}

\makeatletter
\newcommand{\linebreakand}{%
\end{@IEEEauthorhalign}
\hfill\mbox{}\par
\mbox{}\hfill\begin{@IEEEauthorhalign}
}
\makeatother


\author{\IEEEauthorblockN{1\textsuperscript{st} Sandipan Das}
	\IEEEauthorblockA{\textit{Information Science and Engineering} \\
		\textit{KTH Royal Institute of Technology}\\
		Stockholm, Sweden \\
		sandipan@kth.se}
	\and
	\IEEEauthorblockN{2\textsuperscript{nd} Alireza M. Javid}
	\IEEEauthorblockA{\textit{Information Science and Engineering
		} \\
		\textit{KTH Royal Institute of Technology}\\
		Stockholm, Sweden \\
		almj@kth.se}
	\and
	\IEEEauthorblockN{3\textsuperscript{rd} Prakash Borpatra Gohain}
	\IEEEauthorblockA{\textit{Information Science and Engineering
		} \\
		\textit{KTH Royal Institute of Technology}\\
		Stockholm, Sweden \\
		pbg@kth.se}
	
	\linebreakand 
	
	\IEEEauthorblockN{4\textsuperscript{th} Yonina C. Eldar}
	\IEEEauthorblockA{\textit{Mathematics and Computer Science} \\
		\textit{Weizmann Institute of Science}\\
		Rehovot, Israel \\
		yonina.eldar@weizmann.ac.il}
	\and
	\IEEEauthorblockN{5\textsuperscript{th} Saikat Chatterjee}
	\IEEEauthorblockA{\textit{Information Science and Engineering
		} \\
		\textit{KTH Royal Institute of Technology}\\
		Stockholm, Sweden \\
		sach@kth.se}
}
	
\maketitle

\begin{abstract}
	We propose a greedy algorithm to select $N$ important features among $P$ input features for a non-linear prediction problem. The features are selected one by one sequentially, in an iterative loss minimization procedure. We use neural networks as predictors in the algorithm to compute the loss and hence, we refer to our method as neural greedy pursuit (NGP). NGP is efficient in selecting $N$ features when $N \ll P$, and it provides a notion of feature importance in a descending order following the sequential selection procedure. We experimentally show that NGP provides better performance than several feature selection methods such as DeepLIFT and Drop-one-out loss. In addition, we experimentally show a phase transition behavior in which perfect selection of all $N$ features without false positives is possible when the training data size exceeds a threshold. 
\end{abstract}


\section{Introduction}
\label{section:introduction}

Feature selection helps to identify a relevant subset of features from a set of available features. 
It reduces data dimension \cite{dim_1979}, computation complexity and effort in data collection, while improving interpretation of features' role for inference tasks (prediction, classification). 

When using neural networks as a non-linear predictor, a standard methodology of feature selection is through elimination and retraining - a top-down approach. Examples of this approach are Drop-one-out loss \cite{Mao2018} and remove and retrain (ROAR) \cite{ROAR2019}. In these methods, a feature selection algorithm starts with $P$ features in the beginning and then eliminates $(P-N)$ features one-by-one or group-wise to select $N$ important features at the end. In each elimination step, the algorithm performs retraining of the corresponding predictor to find the least important feature to remove. There are two major disadvantages of a top-down approach when $P$ is large. (a) The feature search complexity of the method is $\mathcal{O}((P-N)P) \approx \mathcal{O}(P^2)$ in the regime $N \ll P$. (b) To select $N$ features and the corresponding predictor, we need to start with a predictor that uses $P$ features. Consequently, the predictor would need more training data to account for higher dimensional feature space.

In this article, we address the aforementioned disadvantages. Our contribution is to develop a general methodology for feature selection in neural networks where a bottom-up approach is used - addition and retraining. In this approach, referred to as neural greedy pursuit (NGP), important features are selected one-by-one or group-wise in a sequence until $N$ features are selected. The approach has two major advantages. (a) The feature search complexity is $\mathcal{O}(NP) \ll \mathcal{O}(P^2)$ in the regime $N \ll P$. (b) There is no need to design a predictor that uses more than $N$ features.

\begin{figure*}[t!]
	\centering
	\includegraphics[width=1.0\linewidth, trim = 10 15 0 15,clip]{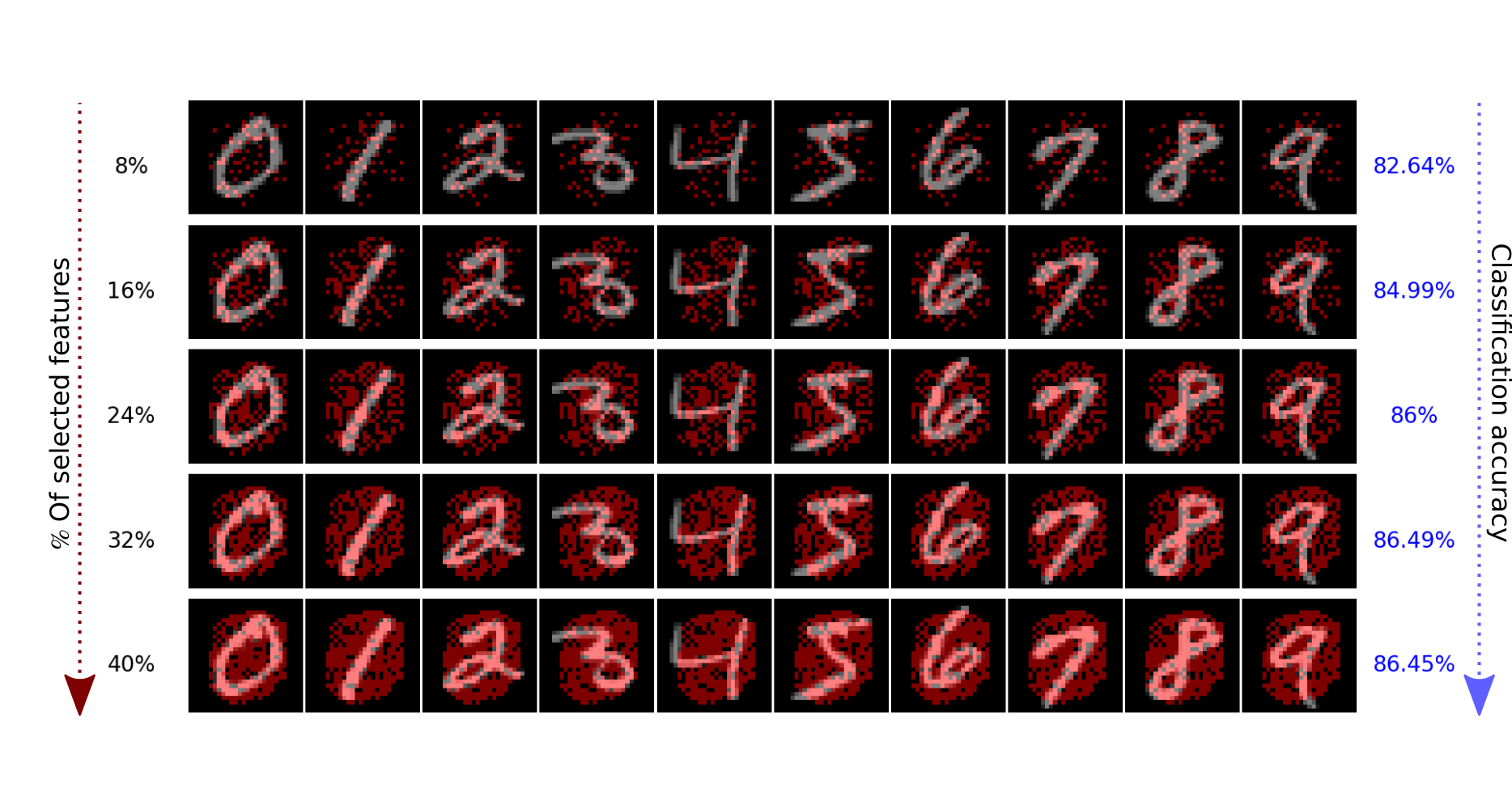}
	\vskip -0.1in
	\caption{Examples of best features (in $\%$) selected by NGP.}
	\label{fig:MNIST_examples}
\end{figure*}

To provide non-linear prediction and exploitation of sparsity when $N \ll P$, NGP provides an appropriate combination of neural networks, and sequential greedy pursuit algorithms for sparse representations \cite{compressed_sensing, eldar_kutyniok_2012}. Note the following point - sparse representation problems are generally unsupervised and consider a linear system model, while our feature selection problem deals with a supervised learning setup and a non-linear system model. NGP is conceptually close to the greedy pursuit algorithms for sparse representations, such as matching pursuit (MP), orthogonal MP (OMP) \cite{MP1993}, and orthogonal least-squares (OLS) \cite{chen1989orthogonal}, where signal components of a sparse signal are selected one-by-one as relevant features.

While the proposed NGP algorithm can accommodate many types of non-linear predictors, we only consider neural networks in this article. We evaluate NGP for various artificial and real datasets, and compare with several methods including least absolute shrinkage selection operator (LASSO) \cite{LASSO1996}, random forest (RF) \cite{rf_2001}, Bayesian additive regression trees (BART) \cite{BART_Chipman_2010}, Drop-one-out loss \cite{Mao2018} and DeepLIFT \cite{deeplift_avanati_2017} with extracted global feature importance via SHAP values \cite{NIPS2017_7062}. In our experiments, we show the following results.
\begin{itemize}
	\item A \emph{phase transition} behavior from an imperfect selection of features to perfect selection when the number of samples in a training dataset exceeds a threshold. To the best of our knowledge, this is the first work to show such behavior where neural networks are used as predictors. We mention that phase transitions occur for sparse representations \cite{donoho2005sparse} and constraint satisfaction problems \cite{saitta2011phase}.
	\item NGP provides a notion of feature importance in a descending order that corresponds to the minimum validation loss. 
	Fig. \ref{fig:MNIST_examples} shows the selected features by NGP for ten different images of the MNIST dataset. Red pixels denote the selected features.
	\item NGP provides $58\%$ better false-positive-selection-rate (FPSR) compared to the top-down approach Drop-one-out loss \cite{Mao2018} for their benchmark artificial dataset while showing consistent and competitive performance across several other datasets.
\end{itemize}
We also mention that NGP is a global feature selection method where features are selected across the population of all samples. There exist local feature selection methods where individual samples are treated \cite{local_global_2018_Adadi}. The population-wise feature selection methods (global) find out an optimal feature subset collectively for all the samples in the population and thus provides global interpretability of the model.

\section{Neural Greedy Pursuit}

\subsection{Problem Formulation}
Let $\mathbf{x} = [x_1 \, x_2 \, \hdots x_P]^{\top} \in \mathbb{R}^{P}$ denote an input data vector (feature vector) where $x_i$ denotes the $i$'th feature, and $\mathbf{t}$ denotes a $Q$-dimensional target vector which we predict using $\mathbf{x}$. Let $\mathcal{S} \subseteq \{1,2, \dots, P \}$, and $\mathbf{x}^{\mathcal{S}}$ represent the part of $\mathbf{x}$ where the components in $\mathbf{x}^{\mathcal{S}}$ are indexed by the elements of $\mathcal{S}$. For example, if $\mathcal{S} = \{2, 5, 9 \}$ then $\mathbf{x}^{\mathcal{S}} = [x_2 \, x_5 \, x_9]^{\top}$. Selection of at most $N$ features to predict the target $\mathbf{t}$ requires to identify $\mathcal{S}$ with $|\mathcal{S}| \leq N < P$. In practice, $ N \ll P$ for most feature selection problems.

Assume that the prediction of target $\mathbf{t}$ is performed using a predictor $\mathbf{y} = f_{\theta}(\mathbf{x})$, where $\mathbf{y}$ is $Q$-dimensional and $\theta$ represents the parameters of the predictor. Examples of such predictors are neural networks, kernel regression and random forests. In a supervised learning setup, we have a training dataset $\mathcal{D}=\{(\mathbf{x}_j,\mathbf{t}_j)\}_{j=1}^J$ with $J$ samples of data-and-target pairs $(\mathbf{x},\mathbf{t})$. Then the problem of jointly learning the parameters of the predictor and selecting at most $N$ features is 
\begin{align}
	\{\hat{\mathbf{\theta}}, \hat{\mathcal{S}} \} \in \underset{\mathbf{\theta},\mathcal{S} }{ \arg\min } \,\, \sum_{j=1}^{J} \mathcal{L}(\mathbf{t}_j,f_{\mathbf{\theta}}(\mathbf{x}^{\mathcal{S}}_j)) + \mathcal{R}(\mathbf{\theta}),
	\label{eq:ANN_minimization}
\end{align}
\noindent such that $|\mathcal{S}| \leq N < P$, where $|\cdot|$ denotes cardinality. 
Here, $\mathcal{L}$ is a chosen loss function, and $\mathcal{R}$ is a regularization term to avoid overfitting, e.g., a simple $\ell_2$-norm weight decay. Examples of loss functions are cross-entropy loss, mean-square loss and hinge loss. The optimization problem \eqref{eq:ANN_minimization} is combinatorial due to search in $\mathcal{S}$ and non-convex to find $\theta$ for neural network-based predictors.

\subsection{Proposed algorithm} \label{ngp_algo}

To address the optimization problem \eqref{eq:ANN_minimization}, we propose a neural greedy pursuit (NGP) algorithm. Let $\mathcal{A} = \{1,\cdots,P\}$. Given $\mathcal{S} \subseteq \mathcal{A}$, the learning of parameters of the corresponding predictor can be shown as the following optimization problem:
\vspace{-8pt}
\begin{align}
	\hat{\mathbf{\theta}}^{\mathcal{S}} \in \underset{\mathbf{\theta} }{ \arg\min } \,\, \sum_{j=1}^{J} \mathcal{L}(\mathbf{t}_j,f_{\mathbf{\theta}}(\mathbf{x}_j^{\mathcal{S}})) + \mathcal{R}(\mathbf{\theta}),
	\label{eq:ANN_minimization_S}
\end{align}
where $\hat{\mathbf{\theta}}^{\mathcal{S}}$ is the optimized parameter. In NGP, we use neural network as the predictor $f_{\mathbf{\theta}}(\cdot)$. 
Cross-validation is applied to optimize the parameters using a validation dataset $\mathcal{D}^{\prime}=\{(\mathbf{x}_j,\mathbf{t}_j)\}_{j=1}^{J^{\prime}}$ and the validation loss,
\vspace{-15pt}
\begin{align}
	\mathcal{L}^{\mathcal{S}} =  \sum_{j=1}^{J^{\prime}} \mathcal{L}(\mathbf{t}_j,f_{\hat{\mathbf{\theta}}^{\mathcal{S}}}(\mathbf{x}_j^{\mathcal{S}})).
	\label{eq:test_loss_S}
\end{align}
Using \eqref{eq:ANN_minimization_S} and \eqref{eq:test_loss_S}, NGP is outlined in Algorithm \ref{alg:NMP}. The algorithm finds features one by one greedily. The indices of selected features are incorporated sequentially in $\mathcal{S}$. This iterative sequential selection process has a high resemblance with variable selection in some prominent sequential greedy pursuit algorithms for sparse representations, such as MP, OMP and OLS \cite{pati1993orthogonal, chen1989orthogonal}. Naturally, the NGP algorithm is efficient for $N \ll P$.
\setlength{\textfloatsep}{3pt}
\begin{algorithm}[h!]
	\caption{: Neural Greedy Pursuit (NGP)}
	\label{alg:NMP}
	\mbox{Input:}
	\begin{algorithmic}[1]
		\STATE Training dataset $\mathcal{D}$ and validation dataset $\mathcal{D}^{\prime}$
		\STATE The maximum number of selected features $N$, and/or a stopping threshold $\eta$
	\end{algorithmic}
	\mbox{Initialization: set $\mathcal{S}_0 \leftarrow \varnothing$, $\mathcal{A} \leftarrow \{1,\cdots,P\}$, iteration $k \leftarrow 0$}

	\begin{algorithmic}[1]
		\REPEAT
		\STATE $k \leftarrow k+1$    \hfill (Iteration counter increment by one)
		\STATE $i_k^{\star} \leftarrow \arg \underset{ i \in \mathcal{A}   } {\min} \,\, \mathcal{L}^{\mathcal{S}_{k-1}^{(i)}}$ where $\mathcal{S}_{k-1}^{(i)} = \mathcal{S}_{k-1} \cup i$  \hfill (Compute loss $\mathcal{L}^{\mathcal{S}_{k-1}^{(i)}}$ using \eqref{eq:ANN_minimization_S} and  \eqref{eq:test_loss_S})
		\STATE $\mathcal{S}_k = \mathcal{S}_{k-1} \cup i_k^{\star}$ \hfill (Greedy choice)
		\STATE $\mathcal{A} \leftarrow \mathcal{A} - i_k^{\star}$.  \hfill (Removing the chosen feature index $i_k^{\star}$ from $\mathcal{A}$)
		\UNTIL{$|\mathcal{S}| > N$} or $\frac{\mathcal{L}^{\mathcal{S}_k} - \mathcal{L}^{\mathcal{S}_{k-1}}}{\mathcal{L}^{\mathcal{S}_k}} < \eta$ \hfill (Stopping condition)
	\end{algorithmic} 	
	\mbox{Output:} 
	\begin{algorithmic}[1]  	
		\STATE Set of indices of selected features: $\hat{\mathcal{S}} = \mathcal{S}_{k-1}$, 
		\STATE Sorted features' indices with descending importance: $i_1^{\star}, i_2^{\star}, i_3^{\star}, \hdots$
	\end{algorithmic}
\end{algorithm}

\textbf{Flexibility to accommodate a combination of predictors:}
In the algorithm, we learn the predictor $f_{\hat{\mathbf{\theta}}^{\mathcal{S}_k}}(\mathbf{x}^{\mathcal{S}_k})$ for each iteration $k$. The predictors $\{ f_{\hat{\mathbf{\theta}}^{\mathcal{S}_k}}(\mathbf{x}^{\mathcal{S}_k}) \}$ can have different architectures and/or types across iterations. For example, $f_{\hat{\mathbf{\theta}}^{\mathcal{S}_k}}(\mathbf{x}^{\mathcal{S}_k})$ can be a neural network, but $f_{\hat{\mathbf{\theta}}^{\mathcal{S}_{k-1}}}(\mathbf{x}^{\mathcal{S}_{k-1}})$ can be a kernel substitution-based predictor such as support vector machine (SVM). It is also possible that, for an iteration, we use a set of different types of predictors, and choose the predictor from the set that provides best loss minimization performance. For example, in iteration $k$, we use both types of predictors, a neural network and a kernel substitution, and then choose the best predictor out of two. We do not perform studies on using different predictors for NGP in this paper, and restrict ourselves to neural networks as predictors.

\textbf{A notion of feature importance:}
An inherent advantage of the sequential choice of features is to provide a notion of feature importance in a sorted manner (descending order). In the first iteration, the algorithm estimates the most important feature. After that, the algorithm estimates the second most important feature and continues. The order corresponds to the effect on the minimization of loss.

\textbf{Phase transition:}
A natural question is whether the NGP algorithm can select all the $N$ important features perfectly in $N$ iterations? Perfect selection depends on the efficiency of predictors $\{ f_{\hat{\mathbf{\theta}}^{\mathcal{S}_k}}(\mathbf{x}^{\mathcal{S}_k}) \}$ and training data size $J$. We will show experimentally that it is possible to achieve perfect selection $\hat{\mathcal{S}} = \mathcal{S} $ when $J$ exceeds a threshold. This is a phase transition behavior that resembles similar behavior for sparse signal recovery problems in sparse representations where perfect signal reconstruction happens when measurements exceed a certain threshold \cite{donoho2005sparse}.

\textbf{Technical limitation:}
Once a feature is selected, it can not be removed in the sequential process. A feature selected in a past iteration can be a false positive or it may no longer remain relevant if another important feature is selected in a current/future iteration. 

\section{Experiments and Discussion} \label{experiments}
In this section, we evaluate NGP on various artificial and real datasets and compare it with a few prominent feature selection methods. The corresponding source codes along with more illustrative simulations are available in the supplementary materials. All the experiments are implemented using a standard core-i7 laptop with 16GB RAM. 
\subsection{Datasets, Performance Measures and Competing Methods} \label{datasets_performance_competing}
{\bf{Datasets:}} For artificial data, we use three well-known physical laws to generate data and a complex non-linear generative data model of \cite{Mao2018}. For real data, we one regression dataset - BOSTON \cite{harrison1978hedonic}, and one classification datatset - MNIST \cite{lecun1998gradient}. 

Ohm's, Planck's and Gravitational laws are used as the three physical laws to generate artificial data. Ohm's law: current $I = \frac{V}{R}$, where $V$ is the voltage across a resistor $R$. Planck's law: the spectral radiance for frequency $\nu$ at absolute temperature $T$ is given by $B = 2 \nu^3 \frac{1}{e^{\frac{\nu}{T}}-1}$. Gravitational law: for two masses $m_1$ and $m_2$ with a distance of $r$, the force is $F = G \frac{m_1 m_2}{r^2}$, where $G$ is the gravitational constant. 
For each law, we generate $J=1000$ samples where feature vectors are 10-dimensional ($P=10$) and the first two (three) components of a feature vector are used to generate the corresponding target for each physical law. All components of the feature vectors are drawn from a uniform distribution $\mathcal{U}(10,20)$. Note that this is a tough condition because statistics of true features typically differ from irrelevant features in real life, but we consider the same statistics for all ten features. 

In order to have a fair comparison with competitive methods, we consider the same artificial data generation model of Drop-one-out loss \cite{Mao2018} as below:
\begin{multline}
	t \! = \! \frac{10 \sin(x_1 \!\vee\! x_2) \! + \! (x_3 \! \vee \! x_4 \! \vee x_5)^3}{1+(x_1 + x_5)^2} + \\ \sin(0.5 x_3)(1 \!+\! \exp^{x_4-0.5 x_3}) \!+\! x_3^2 \!+\! 2 \sin(x_4) \!+\! 2x_5 \!+\! \epsilon,
	\label{eq:Mao_2018_model}
\end{multline}
where $\epsilon \sim \mathcal{N}(0,1)$. The Artificial dataset consists of 600 observations, 300 of them are used for training and the rest for testing. Features $x_1, \cdots, x_5$ and 495 additional irrelevant features are generated by $x_i^{(j)} = \frac{e^{(j)}+z_i^{(j)}}{2}$, $i=1,\cdots,500$, $j=1,\cdots,600$, where $e^{(j)}$ and $z_i^{(j)}$ are independently generated from $\mathcal{N}(0,1)$. Therefore, all of the input features are mutually correlated. 

The BOSTON dataset consists of 339 training samples and 167 test samples. The feature components are $x_1, \dots x_{13}$ and 100 additional features are generated independently from $\mathcal{U}(10,10)$ and appended to the true existing features. 
All the experiments of the MNIST dataset are done using a subset of $J=1000$ samples of MNIST.


{\bf{Performance measures:}} We use fitting mean square error (F-MSE) on traning dataset and prediction mean square error (P-MSE) on test dataset as performance measures. We also show fitting normalized mean square error (F-NME) and prediction normalized mean square error (P-NME) in dB scale. In addtion, we use false positive selection rates $\text{FPSR} = \frac{|\hat{\mathcal{S}} - \mathcal{S}|}{|\hat{\mathcal{S}}|}$ and false negative selection rates $\text{FNSR} = \frac{|\mathcal{S} - \hat{\mathcal{S}}|}{|\mathcal{S}|}$, where $\mathcal{S}$ and $\hat{\mathcal{S}}$ denote the
set of true features and the set of selected features, respectively. 

{\bf{Competing methods:}} We compare NGP with several feature selection methods, including Drop-one-out loss \cite{Mao2018}, RF (number of trees in the forest = 100), LASSO (regression constant = 0.01), BART-50 (50 trees), and GAM (each feature is modeled by a functional form with 5 splines). RF, LASSO and GAM are adapted from scikit-learn package \cite{scikit-learn} while BART is implemented using XBART \cite{pmlr-v89-he19a}. The results of Drop-one-out loss are directly reported from \cite{Mao2018} due to code unavailability. We also compare with a linear correlation-based method where the second-order correlation between $\mathbf{x}$ and $\mathbf{t}$ are calculated, and then the features with the highest correlations are chosen. To compare the behavior of feature selection for image classification (MNIST), we used SOTA instance-based feature importance selector like DeepLIFT. We added the SHAP values of the instances, as DeepLIFT provides local feature importance. DeepLIFT trains a CNN for the MNIST with the following structure: Input - Conv2D(32, 3x3) - MaxPool(2x2) - Conv2D(64, 3x3) - MaxPool(2x2) - Dropout(0.5) - Dense(10) and ran it over 10 epochs with batch size of 32. All the comparisons are evaluated over 10 Monte Carlo simulations.
In the end, for all the competing methods we chose the features having an importance value of greater than 1\%. 

For implementing NGP, we used different neural network architectures such as self-size estimating feed-forward network (SSFN) \cite{SSFN_Saikat}, multilayer perceptron (MLP) \cite{Schmidhuber_2015}, and CNN to show the universality of NGP. We used a single layer SSFN with 100 random neurons, and a single layer MLP with 500 hidden neurons with ReLU activation trained over 10 epochs. The CNN model used in NGP is the same as in DeepLIFT to have a fair comparison across the different competing methods.

{\bf{NN architecture(s) used in our benchmarking:}} We used a simple CNN to show our bottom-up idea as shown in Fig. \ref{fig:CNN_arch}. Hence, we do necessarily outperform the  state-of-the-art approaches. SSFN is also a low complexity algorithm and it does not have high efficiency, in the sense that it did not necessarily produce a high training and testing accuracy when we used 100\% of the features. The SSFN provides around 95\% accuracy using all the features, while state-of-the-art performance for MNIST is more than 99.5\%. The main motivation for us was to train it quickly so that we could observe the behavior of the feature selector algorithm. Instead of our chosen CNN or SSFN, if we used a complex NN that provides high accuracy (such as Efficient-CapsNet), we believe that a substantial performance improvement would have happened and be at the level of state-of-the-art. 
\begin{figure}[!h]
	\centering
	\includegraphics[width=1.0\linewidth, ]{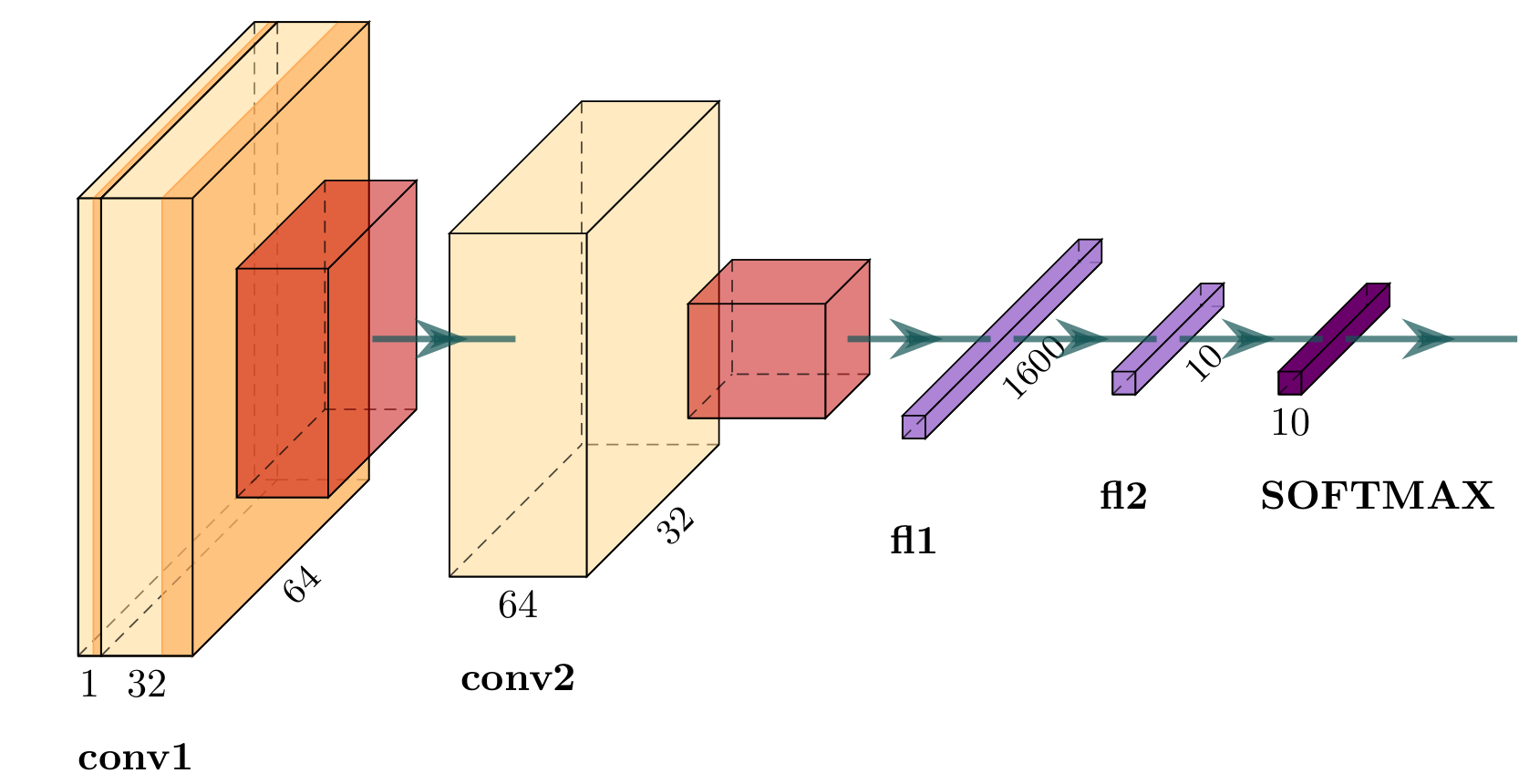}
	\caption{CNN architecture.} \label{fig:CNN_arch}
\end{figure}

\subsection{NGP Performance}
In this subsection, we illustrate various performance behaviors of NGP. Throughout this subsection, NGP uses a SSFN as its predictor $f_{\mathbf{\theta}}(\cdot)$ . The reason for choosing SSFN is its flexibility and low computational complexity in training. That said, NGP is capable to accommodate other neural networks, which will be shown later in Section \ref{Comparison_with_other_methods}. 

Our first hypothesis is that NGP shows improvement in performance as the size of training data increases. There is a threshold in training data size where NGP finds all the relevant features perfectly, which means it shows a \emph{phase transition} behavior. Fig. \ref{fig:Acc_vs_J} shows experimental results using four datasets as the number of training samples increases. Here, $\epsilon=0$ for the artificial data model \eqref{eq:Mao_2018_model}. The plots are shown using 100 Monte Carlo simulations. The decrease in average FNSR with an increase in training data size is shown in Fig. \ref{fig:fnsr}, along with the phase transition behavior in Fig. \ref{fig:phase_change}. Due to a simulation-based study, we define the phase change as $\mathrm{Pr}(\hat{\mathcal{S}} = \mathcal{S}) = 1$ if average FNSR $\leq$ 0.005, otherwise $\mathrm{Pr}(\hat{\mathcal{S}} = \mathcal{S}) = 0$.
Note that $|\mathcal{S}|=N$ is known to NGP for the experiments in Fig. \ref{fig:Acc_vs_J}.

\begin{figure*}[t!]
	\centering
	\begin{multicols}{2}
		\includegraphics[width=1.0\linewidth]{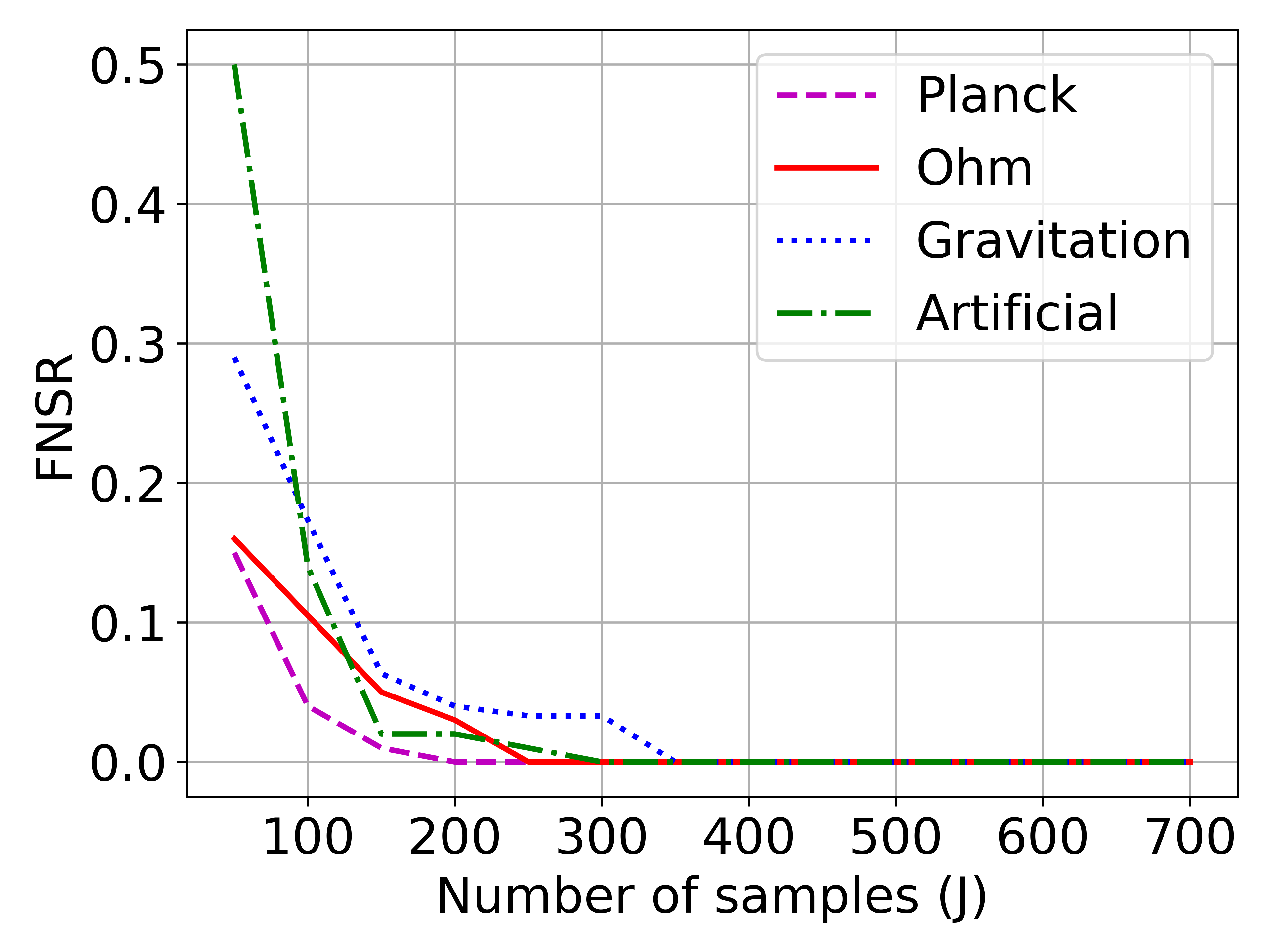}
		\subcaption{Average FNSR versus number of samples.}
		\label{fig:fnsr}
		\includegraphics[width=1.0\linewidth]{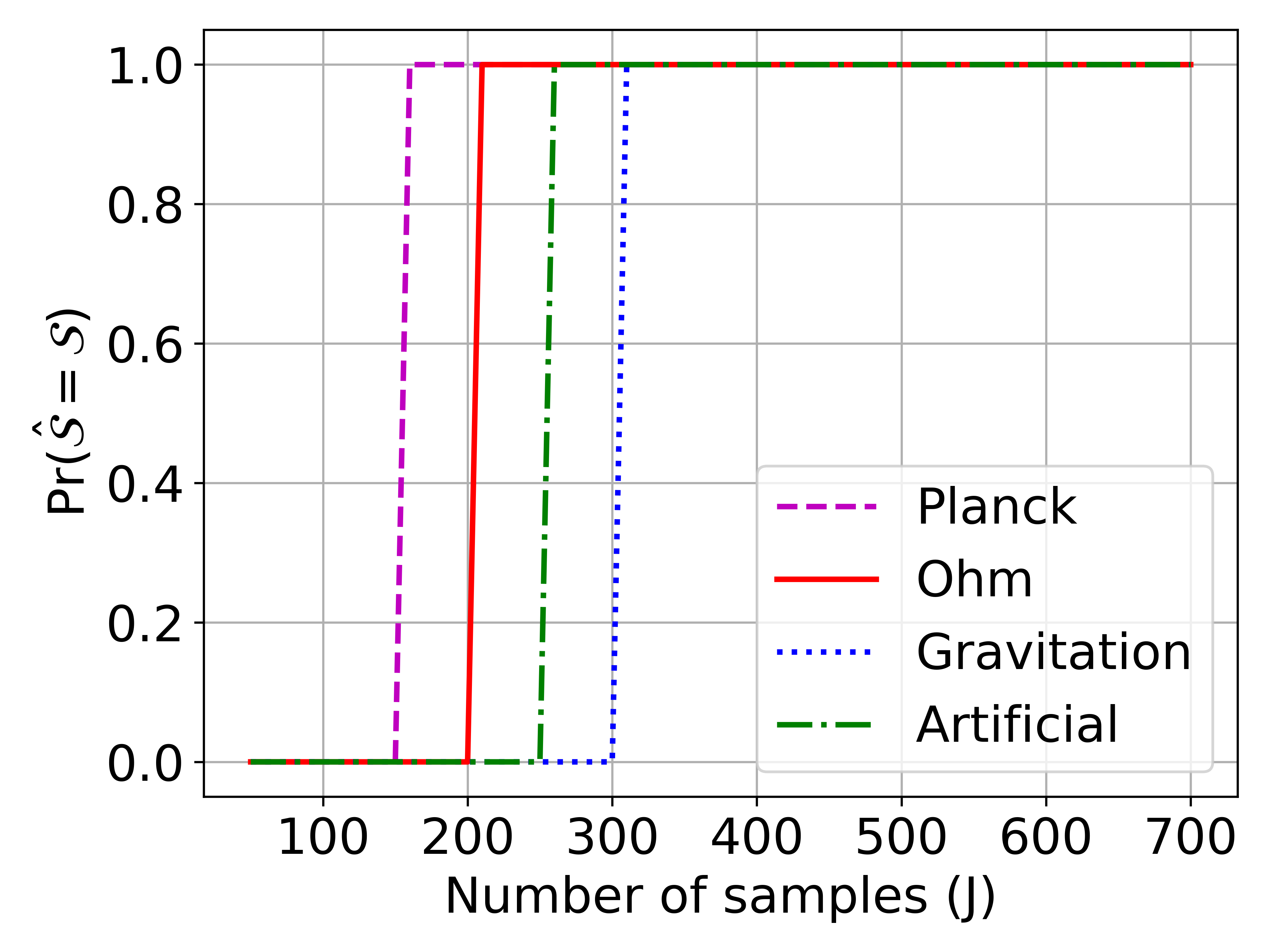}
		\subcaption{Phase transition due to increase in samples.}
		\label{fig:phase_change}
	\end{multicols}
	\caption{Feature selection performance of NGP versus sample size $J$.}
	\label{fig:Acc_vs_J}
\end{figure*}

In practice, we may not know the true cardinality of $\mathcal{S}$. In that case, NGP will continue to add features unless it reaches a stopping criterion. To design a practical criterion, we investigate the behavior of NGP as the number of input features $|\hat{\mathcal{S}}|$ increases in Fig. \ref{fig:Err_vs_feat} for the three physical laws. In all cases, it is seen that the increase in the number of features leads to a sudden change in NME exactly when $|\hat{\mathcal{S}}|$ is around the true cardinality of $\mathcal{S}$. Therefore, it is possible to select the necessary number of features using the changes in NME values and the threshold $\eta$, shown in step 6 of Algorithm \ref{alg:NMP}. 

Finally, let us analyze the behavior of NGP from different aspects on the artificial data generated by \eqref{eq:Mao_2018_model}. In Fig. \ref{fig:Err_vs_feat_artificial}, it is interesting to observe that the sudden change of NME happens where the number of features is equal to five (true cardinality of $\mathcal{S}$ in \eqref{eq:Mao_2018_model}). The performance of NGP in detecting this sudden change depends heavily on the hyperparameter $\eta$ as shown in Fig. \ref{fig:FNSR_vs_eta}. Note that it is possible to achieve FNSR = 0 at the cost of a significantly higher FPSR when $\eta$ decreases. Conversely, it is also possible to achieve FPSR = 0 at the cost of a higher FNSR. Therefore, the choice of $\eta$ is of paramount importance depending on the application of NGP.

While providing low FPSR and FNSR, it is also expected that a feature selection algorithm selects the features according to their relative importance to each other. We demonstrate the aspect of feature importance in Fig. \ref{fig:NGP_sorted_importance}, where $N=5$ is made known to the NGP algorithm. The figure illustrates such characteristics for NGP where we plot the testing NME of NGP against a random selection of features and reversed selection in NGP. As expected, NGP selects the most important feature at first and then continues to find the second most important and so on. Note that the true cardinality of $\mathcal{S}$ is assumed to be known to NGP only in Fig. \ref{fig:NGP_sorted_importance}.

\begin{figure}[t!]
	\centering
	\begin{multicols}{3}
		\includegraphics[width=1.0\linewidth, trim = 10 0 10 0,clip]{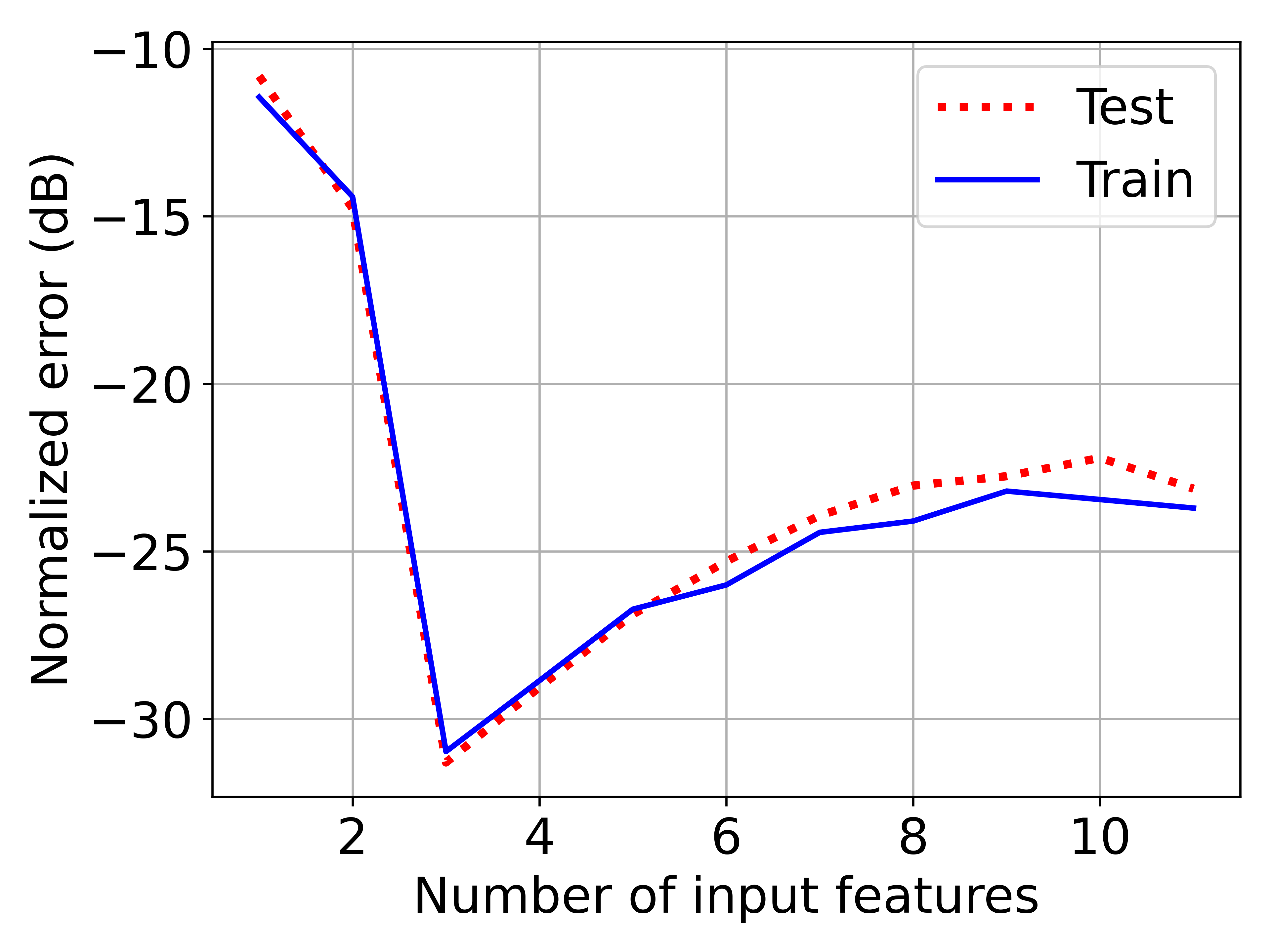}
		\subcaption{Gravitation law}
		\includegraphics[width=1.0\linewidth, trim = 10 0 10 0,clip]{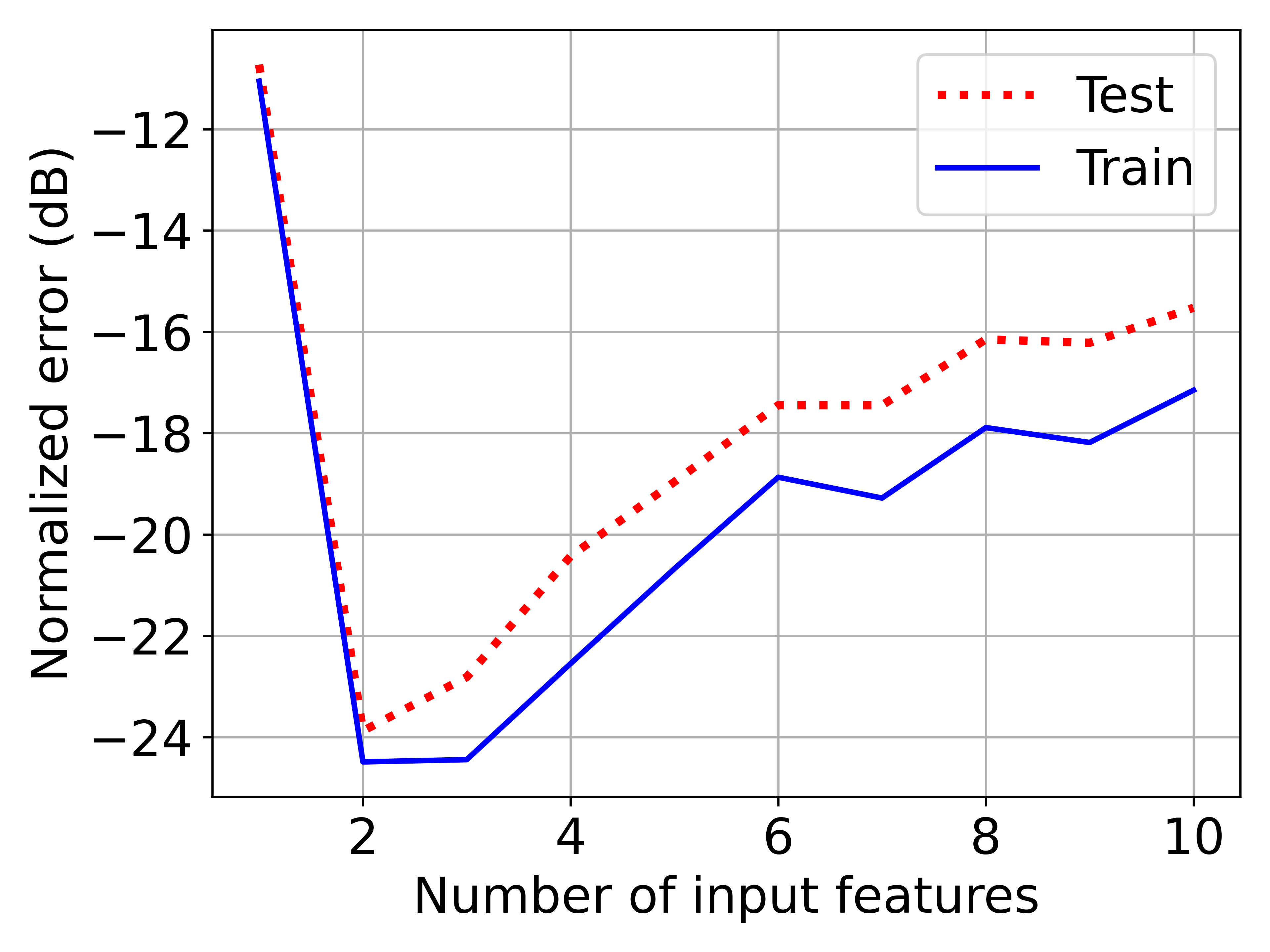}
		\subcaption{Planck law}
		\includegraphics[width=1.0\linewidth, trim = 10 0 10 0,clip]{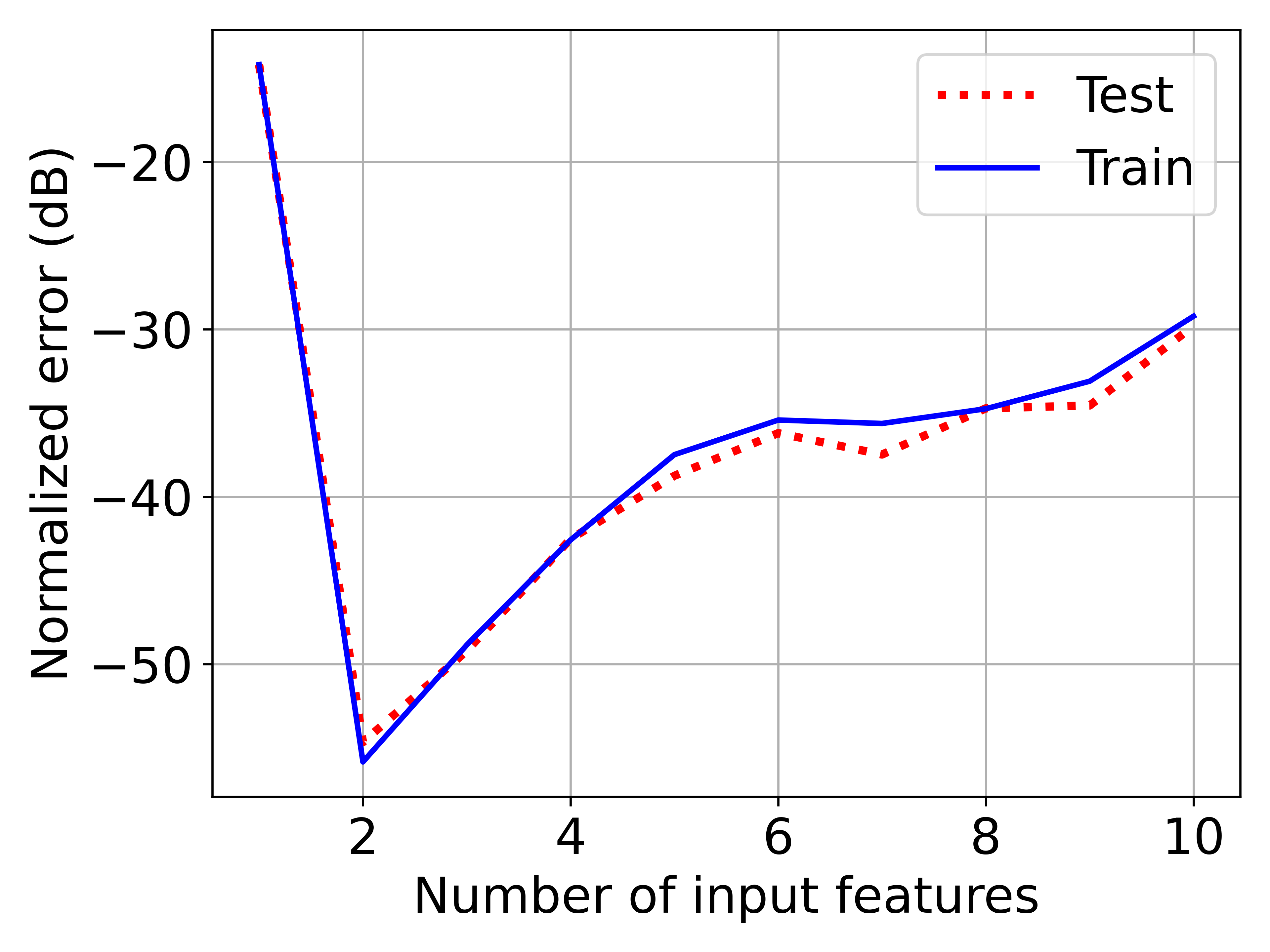}
		\subcaption{Ohm law}
	\end{multicols}
	\caption{Normalized training and testing loss in dB versus the number of input features $|\hat{\mathcal{S}}|$.}
	\label{fig:Err_vs_feat}
\end{figure}

\begin{figure}[t!]
	\centering
	\begin{multicols}{3}
		\includegraphics[width=1.0\linewidth]{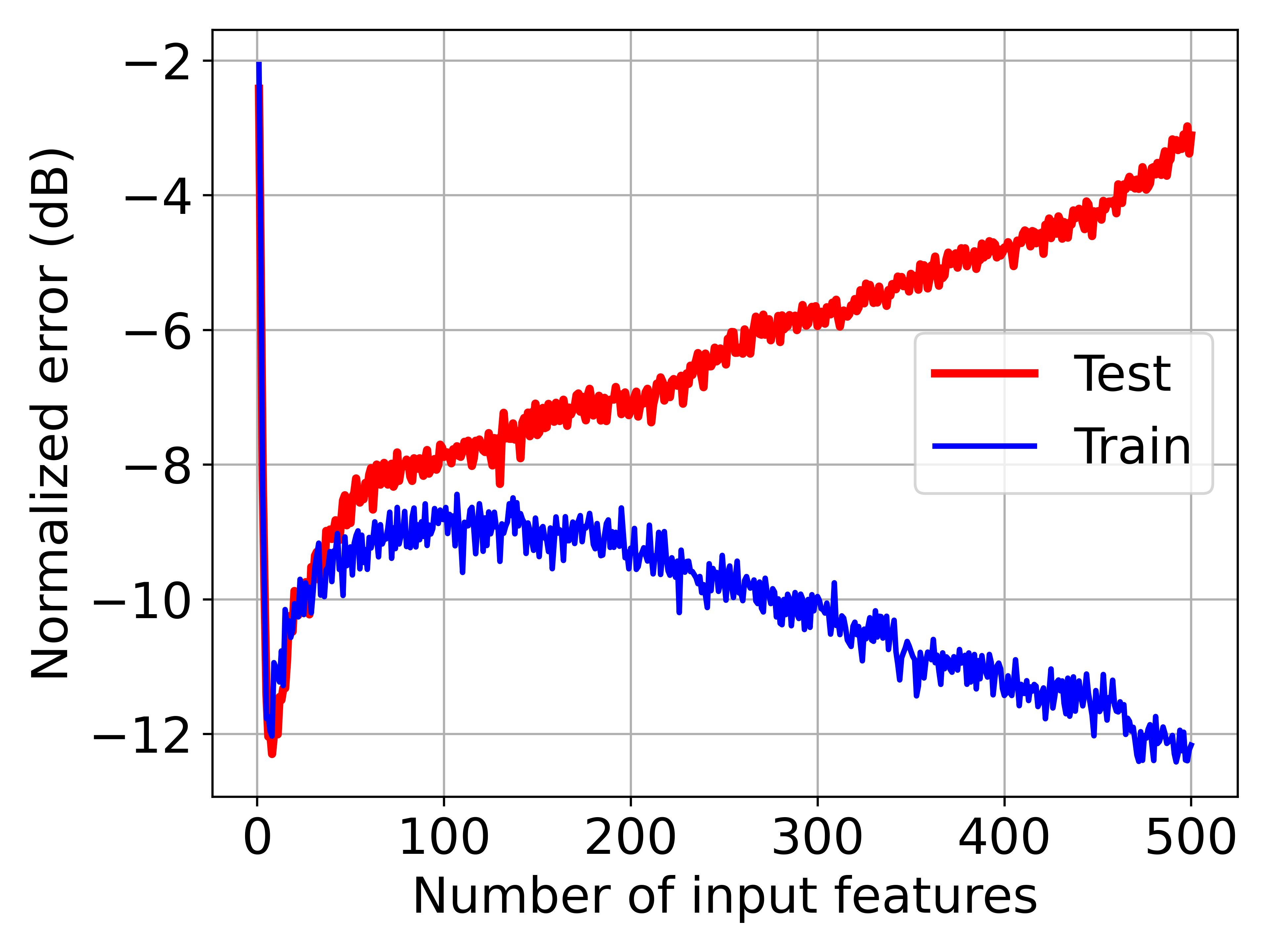}
		\subcaption{Normalized error}
		\label{fig:Err_vs_feat_artificial}
		\includegraphics[width=1.0\linewidth]{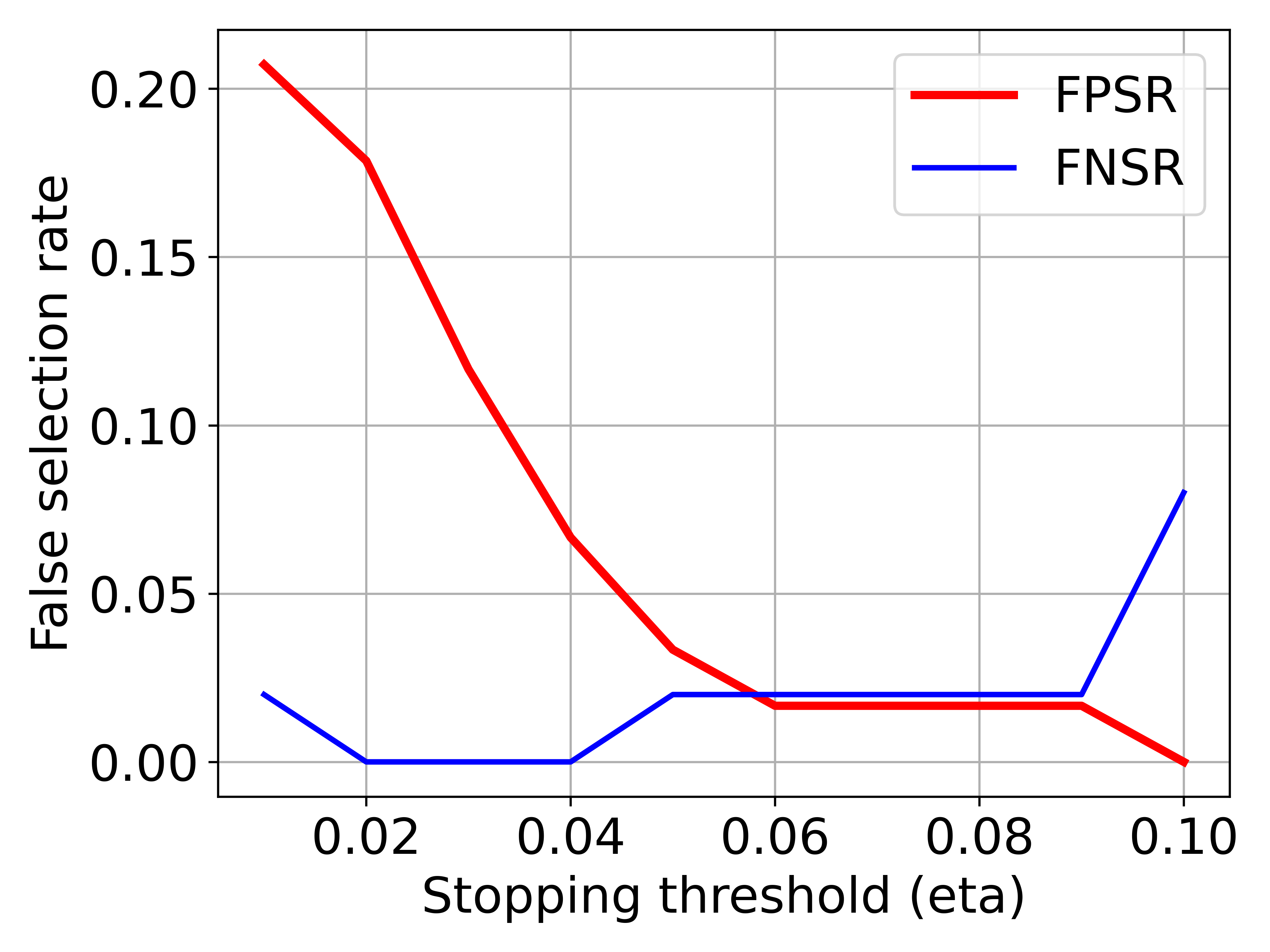}
		\subcaption{False selection rate}
		\label{fig:FNSR_vs_eta}
		\includegraphics[width=1.0\linewidth]{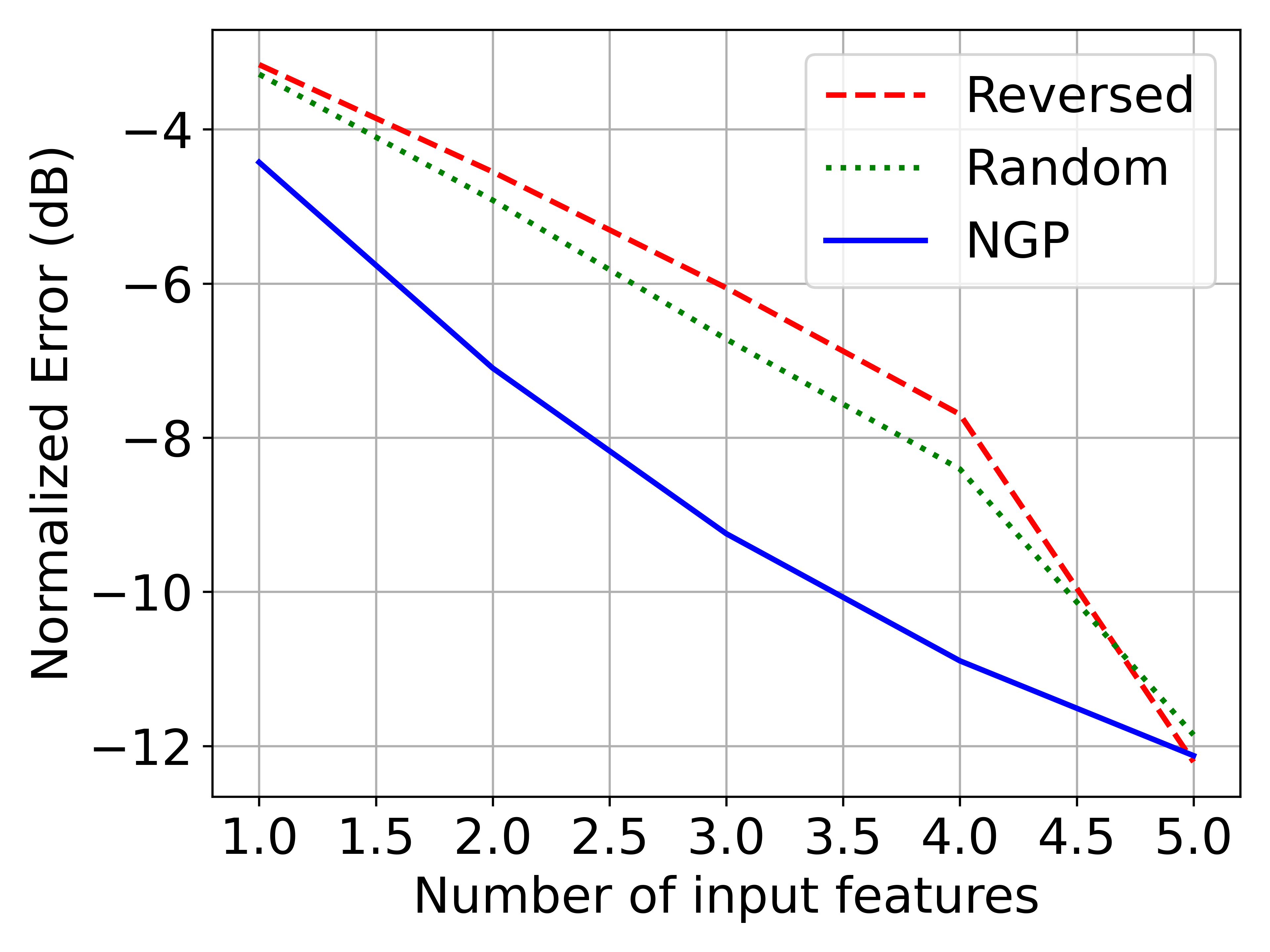}
		\subcaption{Feature rank}
		\label{fig:NGP_sorted_importance}
	\end{multicols}
	\caption{NGP behavior for the artificial data generated by \eqref{eq:Mao_2018_model}.}
	\label{fig:Behavior_NMP}
\end{figure}

\subsection{Comparison with other methods}
\label{Comparison_with_other_methods}

\begin{table}[!hbt]
	\caption{Performance comparison between feature selection methods (averaged over 10 Monte-Carlo simulations). Here the artificial data is according to equation \eqref{eq:Mao_2018_model}.}
	\label{table:Regression_performance}
	\normalsize
	\centering
	\fontsize{14}{12}\selectfont
	\setlength\tabcolsep{0pt}
	\resizebox{\columnwidth}{!}{
		\begin{tabular*}{\textwidth}{@{\extracolsep{\fill}}llcccccc}
			
			\toprule
			& \textbf{Methods} & \textbf{FPSR} & \textbf{FNSR} & \textbf{F-MSE} & \textbf{F-NME} & \textbf{P-MSE} & \textbf{P-NME}\\
			\midrule
			
			\multirow{7}{*}{\rotatebox[origin=c]{90}{\parbox{2 cm}{Artificial data}}} 
			& NGP + SSFN & 0.016 & 0.00 & 0.144 & -11.93 & 0.076 & -12.01 \\
			& NGP + MLP & 0.000 & 0.02 & 0.048 & -10.95 & 0.092 & -10.57 \\
			& Drop-one-out$^{\star}$ & 0.038 & 0.00 & N/A & N/A & N/A & N/A \\
			
			& Correlation & 0.990 & 0.00 & N/A & N/A & N/A & N/A \\
			
			& LASSO & 0.959 & 0.00 & 0.173 & -4.393 & 0.170 & -4.155 \\
			
			& RF & 0.060 & 0.18 & 0.038 & -18.98 & 0.099 & -10.53\\
			
			& BART-50 & 0.991 & 0.00 & 0.114 & -8.305 & 0.132 & -6.655\\
			
			& GAM & 0.985 & 0.00 & 0.147 & -5.841 & 0.155 & -6.167\\
			
			\midrule
			
			\multirow{6}{*}{\rotatebox[origin=c]{90}{\parbox{2.4 cm}{BOSTON dataset}}} 
			& NGP + SSFN & 0.120 & 0.45 & 0.244 & -14.67 & 0.382 & -13.82 \\
			& NGP + MLP & 0.280 & 0.42 & 0.543 & -7.721 & 0.764 & -7.811  \\
			
			& Correlation & 0.884 & 0.00 & N/A & N/A & N/A & N/A \\
			
			& LASSO & 0.717 & 0.29 &  0.289 & -13.20 & 0.467 & -12.09 \\
			
			& RF & 0.000 & 0.55 & 0.086 & -23.71 & 0.345 & -14.72 \\
			
			& BART-50 & 0.892 & 0.07 & 0.614 & -6.658 & 0.889 & -6.494 \\
			
			& GAM & 0.860 & 0.00 & 0.288 & -13.24 & 0.463 & -12.16 \\
			\bottomrule
			\multicolumn{8}{l}{$^{\star}$ \textit{The results are reported from \cite{Mao2018}.} N/A stands for `Not Applicable.'}
		\end{tabular*}
	}
\end{table}

\begin{table}[t!]
	\vskip 0.10in
	\caption{MNIST classification accuracy comparison using top 40\% of the selected features. $J=1000$.}
	\label{table:MNIST_performance_comparison}
	\small 
	\centering
	\fontsize{14}{12}\selectfont
	\setlength\tabcolsep{0pt}
	\resizebox{\columnwidth}{!}{
		\begin{tabular*}{\textwidth}{@{\extracolsep{\fill}}llcc} 
			\toprule
			& \textbf{Methods} & \textbf{Training Accuracy (\%)} & \textbf{Testing Accuracy (\%)} \\
			\midrule
			
			\multirow{6}{*}{\rotatebox[origin=c]{90}{\parbox{1.9cm}{MNIST dataset}}}
			& NGP + SSFN & 93.33 & 86.45  \\
			& NGP + CNN & 86.69 & 86.25 \\
			
			& LASSO & 74.83 & 73.60 \\ 
			
			& RF & 87.26 & 86.64  \\
			
			& BART-50 & 78.27 & 67.52  \\
			
			& SHAP+DeepLift & 70.14 & 68.85\\ 
			
			\bottomrule
		\end{tabular*}
	}
	\vspace{0.29cm}
\end{table}

We start with a comparison of visualizing the selected features for the MNIST image dataset. In addition, we use a limited amount of training data to test robustness against data availability. MNIST has $28\times28$ pixels grey-scale images of hand-written digits. Therefore, we have $P=784$ input features (pixels). The visualization of the selected features out of 784 


\begin{figure*}[!hbt]
	\centering
	\begin{multicols}{2}
		\includegraphics[width=1.0\linewidth, trim = 50 50 50 0,clip]{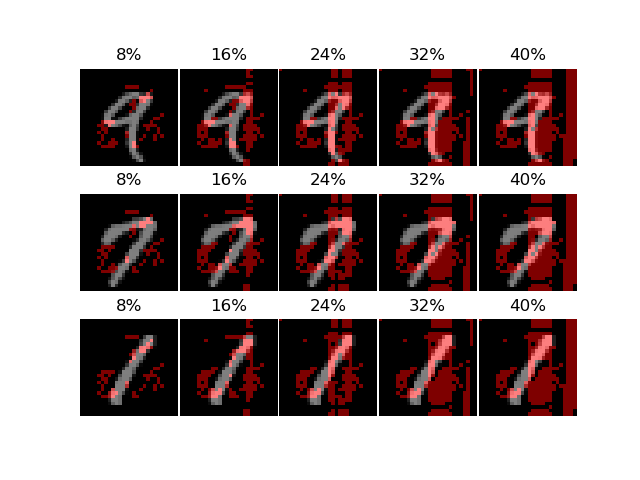}
		\subcaption{LASSO}
		\label{fig:ngp_with_lasso}
		\includegraphics[width=1.0\linewidth, trim = 50 50 50 0,clip]{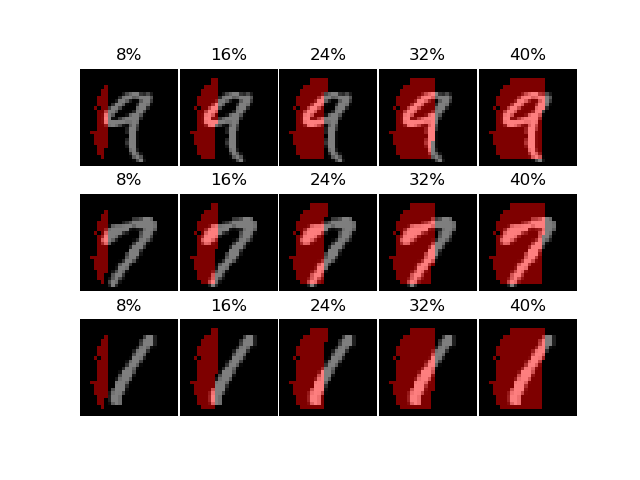}
		\subcaption{RF}
		\label{fig:ngp_with_rf}
		\includegraphics[width=1.0\linewidth, trim = 50 50 50 0,clip]{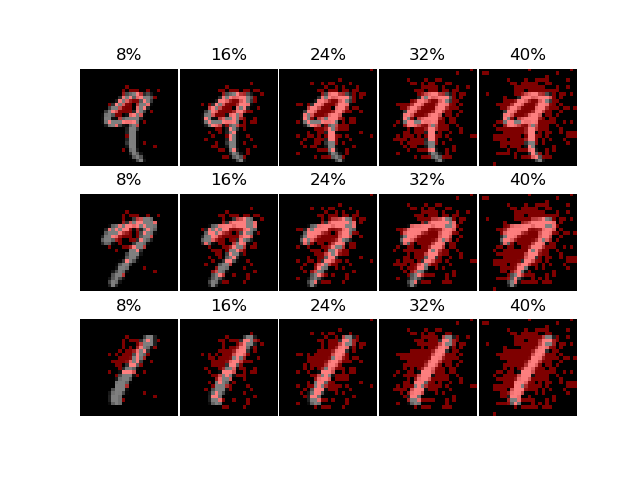}
		\subcaption{BART-50}
		\label{fig:ngp_with_bart}
		\includegraphics[width=1.0\linewidth, trim = 50 50 50 0,clip]{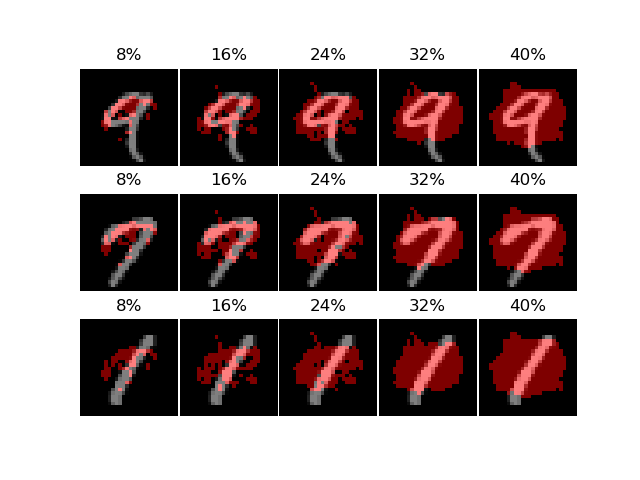}
		\subcaption{SHAP with DeepLIFT}
		\label{fig:ngp_with_deeplift}
		\includegraphics[width=1.0\linewidth, trim = 50 50 50 0,clip]{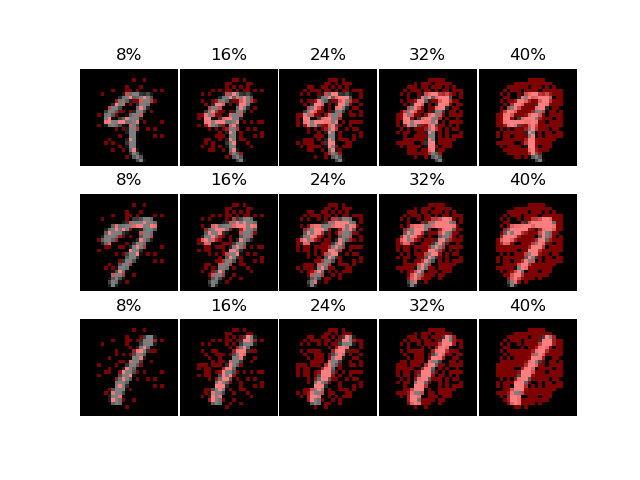}
		\subcaption{NGP, window size = 1$\times$1}
		\label{fig:ngp_with_ssfn_mnist_1_window}
		\includegraphics[width=1.0\linewidth, trim = 50 50 50 0,clip]{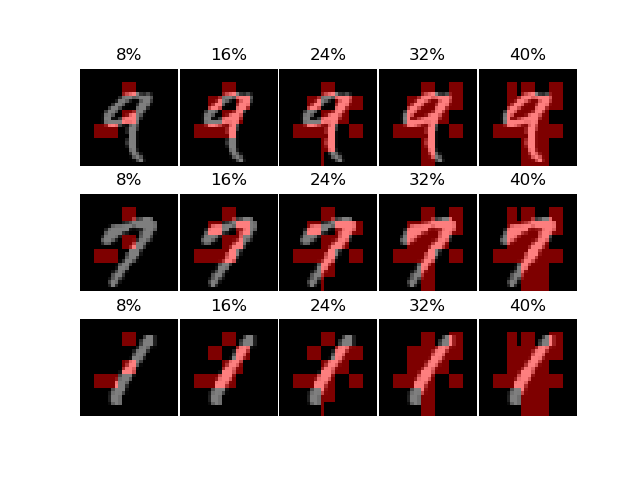}
		\subcaption{NGP, window size = 4$\times$4}
		\label{fig:ngp_with_ssfn_mnist}
	\end{multicols}
	\caption{Examples of \% of best features selected for MNIST.}
	\label{fig:feat_percentage_comparison}
\end{figure*}
\clearpage

\noindent features in percentage is shown in Fig. \ref{fig:feat_percentage_comparison}. We show that a vanilla NGP with window size of 1x1 provides a semantically meaningful feature selection (Fig. \ref{fig:ngp_with_ssfn_mnist_1_window}). It is known that neural networks capture spatial correlations in an image for classification. Therefore, instead of choosing one pixel at a time, we chose a window of 4x4 pixels. Then we greedily select the best feature group by moving the window across the whole image and subsequently select the window which has the lowest MSE loss (Fig. \ref{fig:ngp_with_ssfn_mnist}). We note that the selected features of NGP cover almost all the pixels in the area of digit `9'. The group selection parameter is a tunable hyper-parameter. 

We now show quantitative comparison results in Table \ref{table:Regression_performance} for regression and Table \ref{table:MNIST_performance_comparison} for MNIST classification. From Table \ref{table:Regression_performance}, we observe that NGP provides a competitive performance or significantly better vis-a-vis other methods. For example, let us consider the performance for the artificial data model \eqref{eq:Mao_2018_model} where we note that NGP with SSFN provides $58\%$ FPSR improvement compared to the top-down approach Drop-one-out loss. NGP is found to be significantly better than all the other competing methods including RF.

Next we consider the Table \ref{table:MNIST_performance_comparison} for MNIST digit classification.

\noindent Here, we consider, best $40\%$ feature selection by respective methods for classification accuracy. We use $J=1000$ training samples for feature selection to show robustness against a limited amount of data availability. Predictors are tested using 10000 samples. It is observed that NGP with the best $40\%$ feature selected, provides a classification accuracy similar to RF and significantly better than other methods. 

\section{Conclusion}
We conclude that the sequential feature selection in NGP as a bottom-up approach is efficient in the sense of computation and performance. NGP provides semantically meaningful feature importance, demonstrated for image data in a classification task. We also show a phase transition behavior - $N$ features are perfectly selected in $N$ iterations when the training data size exceeds a threshold. The NGP method can work with different predictors as well as a combination of predictors in each iteration. To explore the generality of NGP, other variants of predictors and their combinations can be considered in the future.  

\section{Acknowledgements}
This research has been conducted as part of development of autonomous transport solutions at Scania. It was jointly funded by Swedish Foundation for Strategic Research (SSF) and Scania. The research was also affiliated with Wallenberg AI, Autonomous Systems and Software Program (WASP). 
\newpage
\bibliographystyle{IEEEbib}
\bibliography{./main}

\end{document}